\documentclass[11pt]{article}
 
\usepackage{dcolumn}
\usepackage{bm}
 
\usepackage{natbib}

\usepackage{subcaption} \usepackage{bbold}
\usepackage{multirow} \usepackage{url}
 
\usepackage[pdftex]{graphicx} \usepackage[pdftex]{color}
 
\newcommand{\secref}[1]{\S\ref{#1}}
\newcommand{\figref}[1]{Figure~\ref{#1}}
\newcommand{\tabref}[1]{Table~\ref{#1}}
\newcommand{\eqref}[1]{Formula~(\ref{#1})} 
\newcommand{\md}{{\em Moby Dick}}
\newcommand{\average}[1]{\ensuremath{\langle#1\rangle} }

\usepackage[ruled,vlined]{algorithm2e}

\begin{document} 
 
\title{A Comparison of Two Fluctuation Analyses\\ for Natural Language
  Clustering Phenomena \\ ---Taylor vs. Ebeling \& Neiman Methods---}
 
\author{Kumiko Tanaka-Ishii \hspace*{1cm} Shuntaro Takahashi
  \\ Research Center for Advanced Science and Technology \\ University
  of Tokyo \\ {\tt \{kumiko, takahashi\}@cl.rcast.u-tokyo.ac.jp}\\ }
 
\date{\today}
 
\maketitle
 
\begin{abstract} 
This article considers the fluctuation analysis methods of Taylor and
Ebeling \& Neiman. While both have been applied to various phenomena
in the statistical mechanics domain, their similarities and
differences have not been clarified. After considering their
analytical aspects, this article presents a large-scale application of
these methods to text. It is found that both methods can distinguish
real text from independently and identically distributed (i.i.d.)
sequences. Furthermore, it is found that the Taylor exponents acquired
from words can roughly distinguish text categories; this is also the
case for Ebeling and Neiman exponents, but to a lesser
extent. Additionally, both methods show some possibility of capturing
script kinds.
\end{abstract} 
 
\section{Introduction} 
Clustering phenomena are frequently observed to underlie complex
systems. In particular, language possesses clustering phenomena that
arise mainly from changes in context.
 
A natural way to quantify the degree of clustering is with {\em
  fluctuation analysis}. Historically, three forms of fluctuation
analysis have been used to study complex systems. While these methods
share many similarities, they have been treated and used in distinct
ways. The first is the Hurst method, a numerical time series method
that was originally used to analyze the flow of the Nile River. Its
application to natural language involves transforming a word sequence
into a numerical sequence of frequency ranks
\citep{montemurro}. Although the idea of applying river flow analysis
to natural language is interesting, the transformation of words into
ranks is somewhat arbitrary. Moreover, the essential problem of the
Hurst method is that it measures the degree of fluctuation by the
difference between the maximum and minimum samples within a segment,
not the variance, causing it to incorporate outliers.
 
Unlike Hurst's method, the second and third forms of fluctuation
analysis exploit the variance. The second form was originally proposed
in \cite{taylor-smith}, where it was used to analyze crops. Taylor
\citep{taylorNature} subsequently used it to analyze biological
colonies. Since then, this method has been referred to as Taylor
analysis, and it has been applied to a wide variety of phenomena
\citep{taylor}. In particular, it has been applied to language
\citep{altmann14-taylor}, notably to large-scale language corpora
\citep{ jpc18, acl18}. In this last case, Taylor's method measures the
variance of every word type in a segment, which is a given-length
subsequence of a text.
 
The third form, generally referred to simply as fluctuation analysis,
measures the overall growth of the variance with respect to the
segment length $l$. A variety of methods can be regarded as
fluctuation analyses. In particular, the detrended fluctuation method
and multifractal methods \citep{Peng, Kantelhardt2002} have been
applied to continuous data. A simple and natural application to
natural language was reported by Ebeling \& Neiman
\citep{Ebeling1995}, which we will refer to as the EN method, or EN
analysis. The EN method was initially intended for characters and used
to study the {\em Bible}.
 
In all three forms of fluctuation analysis, the outcome appears in the
form of a power law. However, except for this similarity, the
commonalities and differences between the above methods have not been
thoroughly examined until now. To address this gap in our
understanding, we decided to compare the Taylor and EN methods
qualitatively and quantitatively.

\section{The Two Methods} 
Suppose that a set of events, $W$, consists of linguistic elements
$w_k \in W$, such as words or characters. Furthermore, let a text be a
sequence of elements, $X = X_1,X_2,\ldots,X_i,\ldots, X_{N}$, where
$X_i=w_k \in W$ for all $i=1, 2, \ldots, N$.
 
For a segment of length $l \in \mathbb{N}$ (a positive integer), $X$
can be statistically analyzed w.r.t. a segment as follows. Let
$c(w_k,l)$ indicate the count of word $w_k$ in the segment, and let
its mean and variance be $\mathbb{E}[c(w_k,l)]$ and
$\mathbb{V}[c(w_k,l)]$, respectively.
 
\subsection{Taylor's Method} 
Taylor's method focuses on the distribution of variances of words
$w_k$ for a given $l$. Precisely speaking, it examines whether the
following relation holds for words $w_k \in W$:
\begin{equation} 
  \mathbb{V}[c(w_k,l)] \propto \mathbb{E}[c(w_k,l)]^{\alpha}.
\end{equation} 
In what follows, we will call the exponent $\alpha$ the Taylor
exponent.
 
\subsection{EN Method} 
The EN method considers the following function $z(l)$, which is the
sum of variances of all elements of $W$:
\begin{eqnarray} 
  z(l)&=&\sum_{w_k\in W} \mathbb{V}[c(w_k,l)]. \label{form:z}
\end{eqnarray} 
The goal of the EN method is to determine whether $z(l)$ shows a
power-law behavior, as follows:
\begin{equation} 
  z(l)\propto l^{\beta}.
\end{equation} 
Hereafter, we will call $\beta$ the EN exponent.

\begin{figure}[t] 
\begin{subfigure}[t]{.49\linewidth} 
\centering \includegraphics[width=60mm]{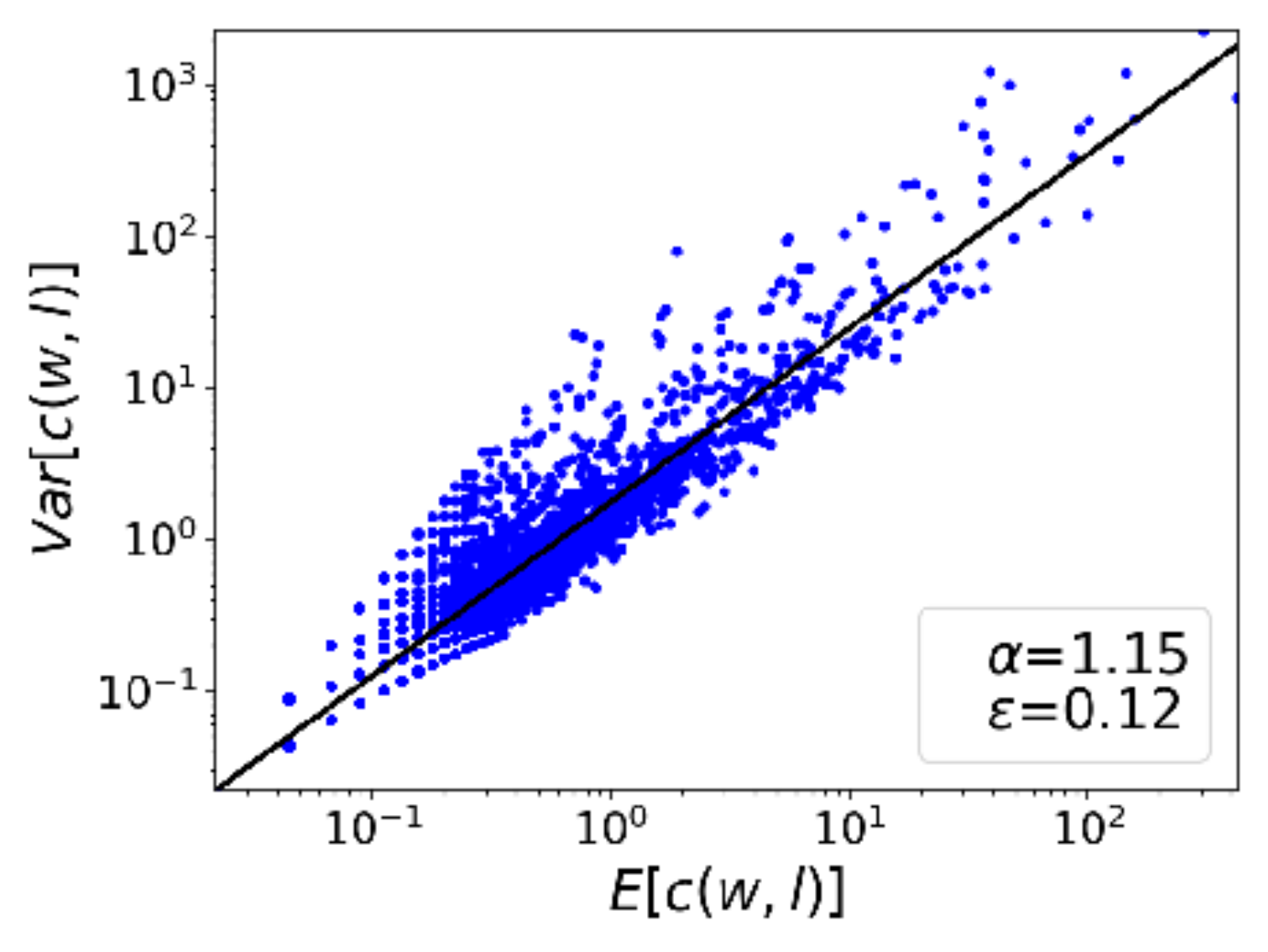}
\caption{Taylor's method}
\end{subfigure} 
\begin{subfigure}[t]{.49\linewidth} 
\centering \includegraphics[width=60mm]{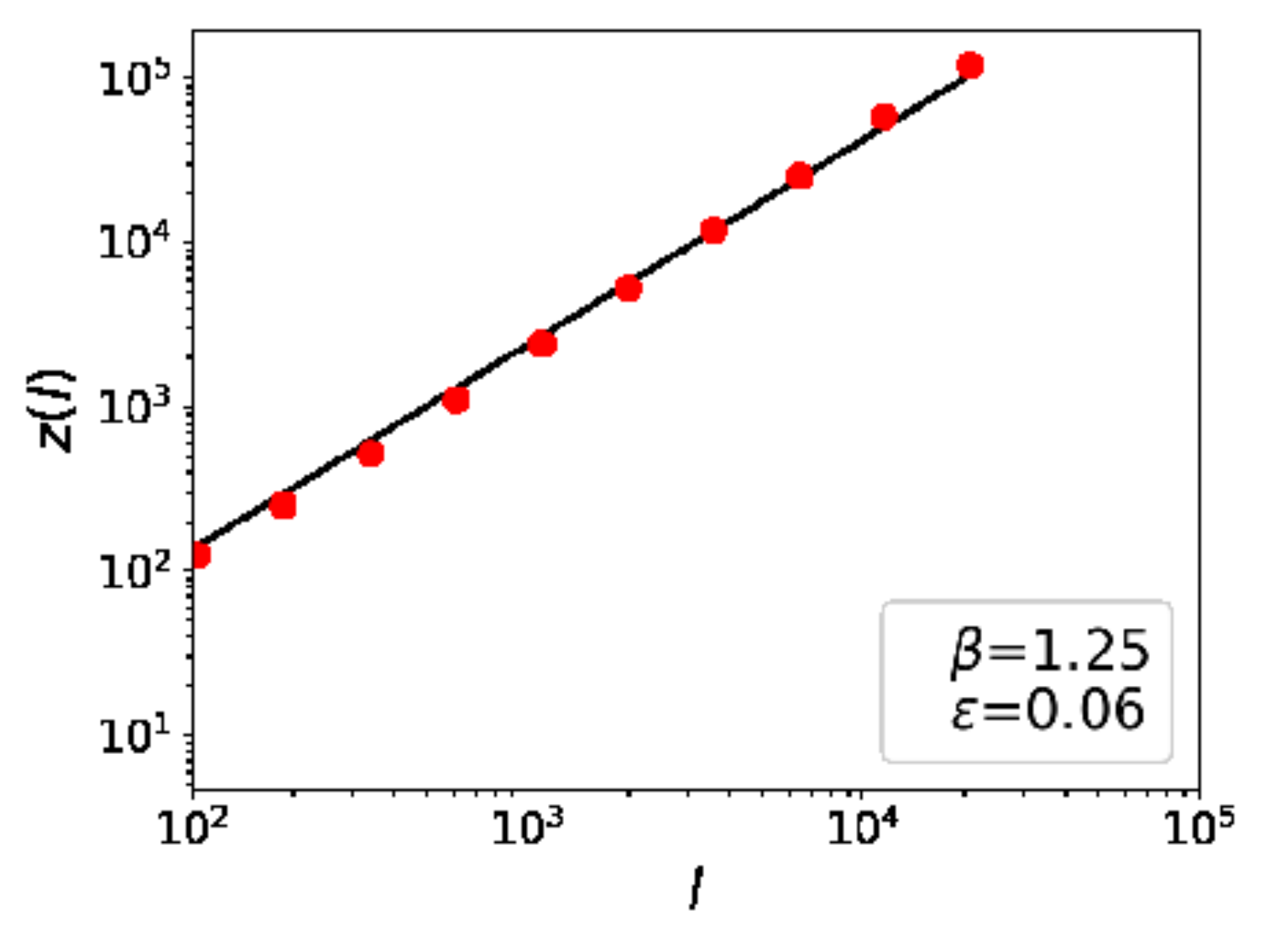}
\caption{EN method}
\end{subfigure}  
\caption{Taylor and EN analyses of {\md}.}
\label{fig:md} 
\end{figure} 

\subsection{Estimation of the Exponent} 
Given a text of length $N$, $\mathbb{E}[c(w_j,l)]$ and
$\mathbb{V}[c(w_j,l)]$ can be determined through the simple procedure
described in Appendix A. In particular, for a sample text, $J$ points
$(x_j,y_j)$ are acquired, where $j=1,\ldots,J$. In the case of
Taylor's method, $x_j \equiv \mathbb{E}[c(w_j,l)]$ and $y_j \equiv
\mathbb{V}[c(w_j,l)]$ for $j=1\ldots,|W|$, whereas in the EN method,
$x_j\equiv l_j$ and $y_j\equiv z(l_j)$ for $0 < l_j < L$,
$j=1,\ldots,J$, with $L$ being some maximum length within a
document of length $N$ (Appendix A gives an example).

We want to fit the points $(x_j,y_j)$ to a power function $y = A \log
x^C$, where $A$ and $C$ are functional parameters, and $C$ is either
$\alpha$ or $\beta$. In log-log coordinates, the power function
becomes a linear plot, and the points can be fitted using the
least-squares method as follows:
\begin{eqnarray} 
  \hat{A},\hat{C} &=& \arg \min_{A,C} \varepsilon(A,C),
  \\ \varepsilon(A,C) &\equiv& \sqrt{\frac{1}{J} \sum_{j=1}^J (\log
    y_j - \log A x^C_j)^2}, \label{eq:error}
\end{eqnarray} 
where $\varepsilon(A,C)$ denotes the square error from the fitted
function, i.e., the residual. That is, the exponent $C$ is determined
by minimizing the residual.

\subsection{Example Analyses} 
\label{sec:example} 
\figref{fig:md} shows the results of Taylor and EN analyses of the
words of the novel {\em Moby Dick} by Melville. Appendix A explains
the procedures used to make these plots, while \tabref{tab:data} of
Appendix B summarizes the statistical details of {\em Moby Dick}. Note
as well that all figures in this article use the color scheme of
\figref{fig:md}: blue for the results of Taylor's method and red for
those of the EN method.
 
In (a), the horizontal axis indicates the mean occurrence in segments,
while the vertical axis indicates the variance. Every point
corresponds to a word type. It can be seen that the data points are
aligned roughly linearly, although vertical deviation is also
present. The exponent is estimated as $\alpha=1.15$, with
$\varepsilon=0.12$.
 
In (b), the horizontal axis indicates the segment length, $l$, while
the vertical axis indicates the logarithm of the variance. Each red
dot represents $z(l)$, the sum of variances of all elements $w$, for a
given segment length $l$. The plot clearly shows a power-law
tendency. The exponent is estimated as $\beta=1.25$, with
$\varepsilon=0.06$.
 
\subsection{Analytical Background} 
\label{sec:theory} 
It is empirically known that $\alpha$ and $\beta$ each have values in
the range $1 \leq \alpha, \beta \leq 2$. In general, $\alpha, \beta
=1$ indicates that the events, in this case words, form a
sequence whose elements are independently and identically distributed
(i.i.d.), whereas both values being $2$ indicates that the events are
clustered.
 
In the former case, for an i.i.d. process, 
the number of occurrences of a word in a
segment of length $l$ obeys a Poisson distribution. Accordingly, the
mean and variance are equal, and trivially, $\alpha = 1$.
Furthermore, because the mean grows linearly with $l$, so does the
variance. Therefore, $\beta = 1$ as well. A formal mathematical
analysis of Taylor's method for the i.i.d. case is given in
\citet{acl18}. The same analysis can be easily extended to the EN
method.
 
In the latter case, when the events are clustered, the exponents
rise. In particular, $\alpha, \beta=2$ when all segments contain the
same proportions of the elements of $W$. For example, suppose that $W
= \{a, b\}$. If $b$ occurs twice as often as $a$ in all segments
(e.g., three $a$ and six $b$ in one segment, two $a$ and four $b$ in
another, etc.), the mean for $b$ is twice that for $a$, and similarly
for the variances. Thus, $\alpha=2$. Furthermore, if 
$l$ is made 
$m$ times larger, then the variances of the occurrences of
$a$ and $b$ each become $m^2$ times larger, and so
$\beta = 2$, as well. This simple example indicates that
co-occurrences of elements have the effect of enlarging the exponents.
 
The values of $\alpha$ and $\beta$ can be analytically analyzed in a
different way through rare events. Consider a rare word that occurs
only a few times in the entire text. If every occurrence is in a
different segment, then it can be analytically shown (see
\cite{jpcadd19}) that $\alpha=1$, whereas if all occurrences are in
the same segment, then $\alpha=2$. A similar argument applies to the
cases of $\beta=1$ and $\beta=2$ in the EN method.

\subsection{Qualitative Comparison of the Two Methods} 
\label{sec:analysis} 
EN and Taylor analyses are each based on $\mathbb{V}[c(w,l)]$. Despite
this similarity, they treat sequences from different perspectives, and
the fluctuations that they capture are different. In both analyses, a
text has a characteristic that fluctuation is amplified by a power
law, but the EN method focuses on $l$ in $c(w,l)$, whereas Taylor's
method focuses on $w$.
 
Let us summarize the differences between the two methods. First,
Taylor's method depends on $l$, whereas the EN method does
not. However, this fact does not make Taylor's method insignificant,
because the overall qualitative understanding it provides has been
empirically shown not to depend on $l$ \citep{taylor,jpc18}.
 
Second, while the EN method is valid even when the size of the set of
elements is small, such as when $W$ is the set of Roman alphabetical
characters, Taylor's method requires a large number of types of
elements to give meaningful results.
 
Taylor's method can be regarded as the first step of the EN method;
i.e., it determines $z(l)$ in \eqref{form:z}. The EN method sums the
variances without regard to the power dependency underlying the
events, and the sum accumulates to $z(l)$ for a particular $l$. This
procedure corresponds to Taylor's method, which produces $\alpha$ by
analyzing the distribution. Therefore, the results of the EN method
are partly a consequence of Taylor's method.
 
With these qualitative differences in mind, let us now compare the two
methods empirically.

\section{Datasets} 
\label{sec:data} 
\tabref{tab:data} in Appendix B lists the datasets used in this
study. The data consist of literary texts, newspaper corpora, and
language-related data from sources such as music and programming
languages. Appendix B explains the details of the data and how they
were preprocessed.

\begin{figure} 
	\begin{subfigure}[t]{.49\linewidth} 
	\centering \includegraphics[width=60mm]{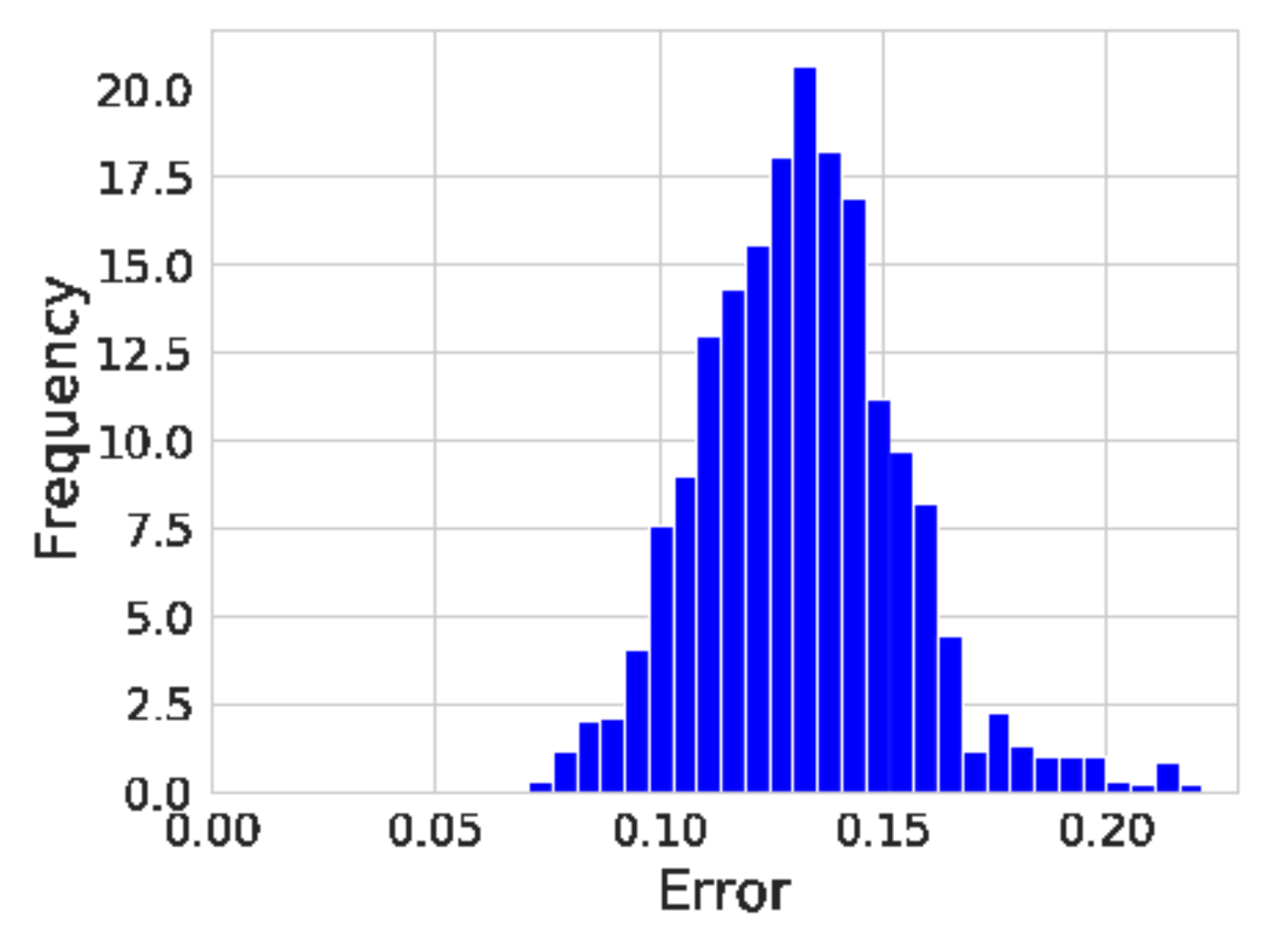}
	\caption{Taylor's method}
	\end{subfigure}  
	\begin{subfigure}[t]{.49\linewidth} 
	\centering \includegraphics[width=60mm]{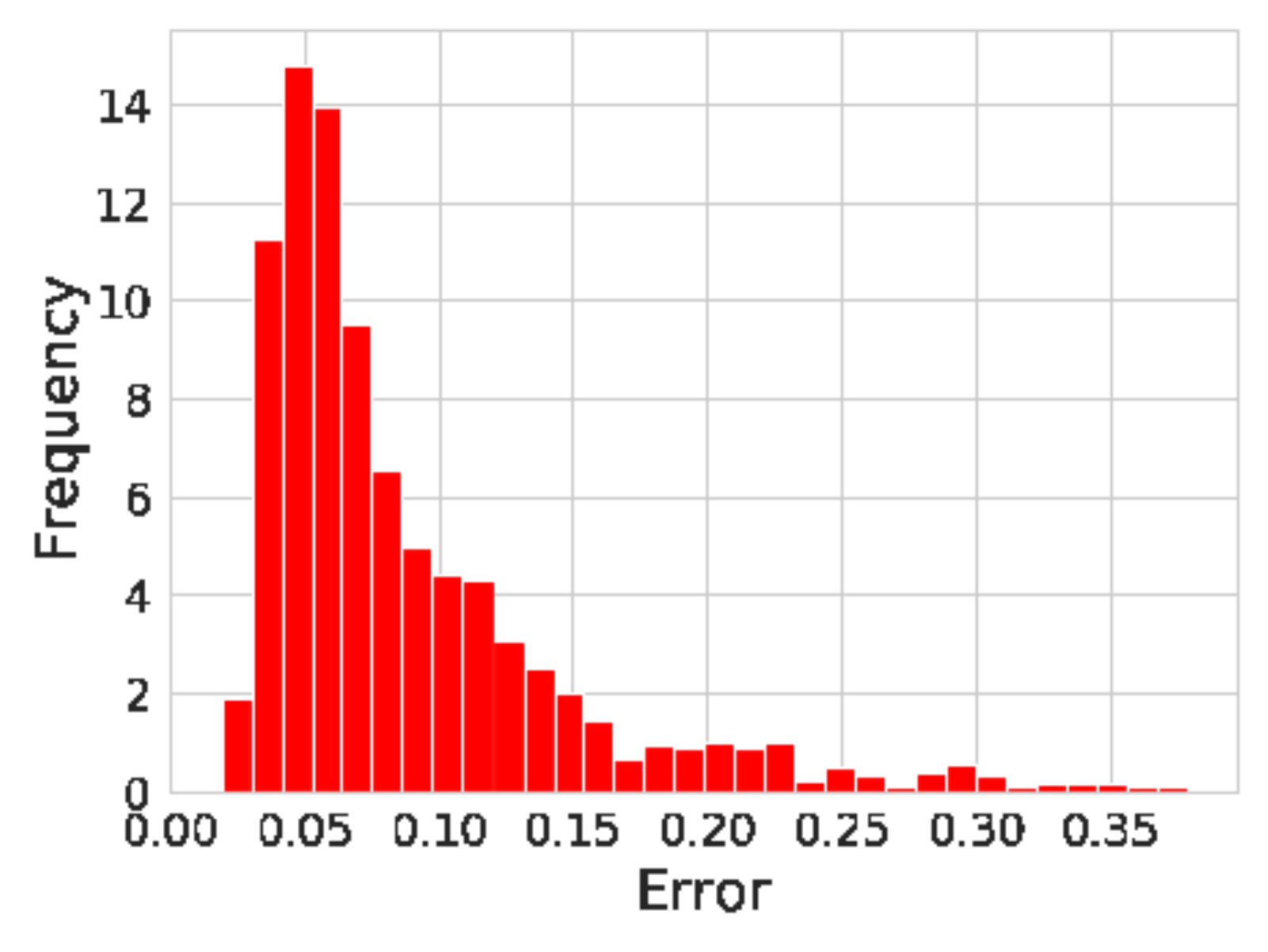}
	\caption{EN method}
	\end{subfigure} 
	\caption{\label{fig:error} Distribution of the fitting error
          for words in long literary texts collected from Project
          Gutenberg. The horizontal axis indicates the $\varepsilon$
          values, while the vertical axis indicates the number of
          texts. $l=5620$ for Taylor's method.}
\end{figure} 

\section{Basic Statistical Properties of the Two Methods} 
\label{sec:basic_properties} 
This section describes the fundamental statistical properties of the
two methods. First, let us consider their fits. \figref{fig:error}
shows the distribution of the fitting error to a power function, i.e.,
the value of $\varepsilon$ as defined in \eqref{eq:error}. The
horizontal axis indicates the error, while the vertical axis indicates
the number of sample texts. Both methods were applied to the words in
the literary texts from Project Gutenberg (\tabref{tab:data}).

For Taylor's method, the fitting error was calculated across a
distribution of data points, where each point represents a word. The
words are distributed as shown in the left graph of \figref{fig:md},
where it is clear that the $\varepsilon$ values tend to be larger in
comparison with those in the right graph for the EN method.

On the other hand, for the EN method, only one data point, i.e.,
$z(l)$, was taken for each value of the segment length $l$. Therefore,
the error is relatively small compared with that of Taylor's
method. The fitting error of the EN method tends to increase
with the estimated exponent value, however, and it is large for some texts.

\begin{figure} 
\centering
\includegraphics[width=80mm]{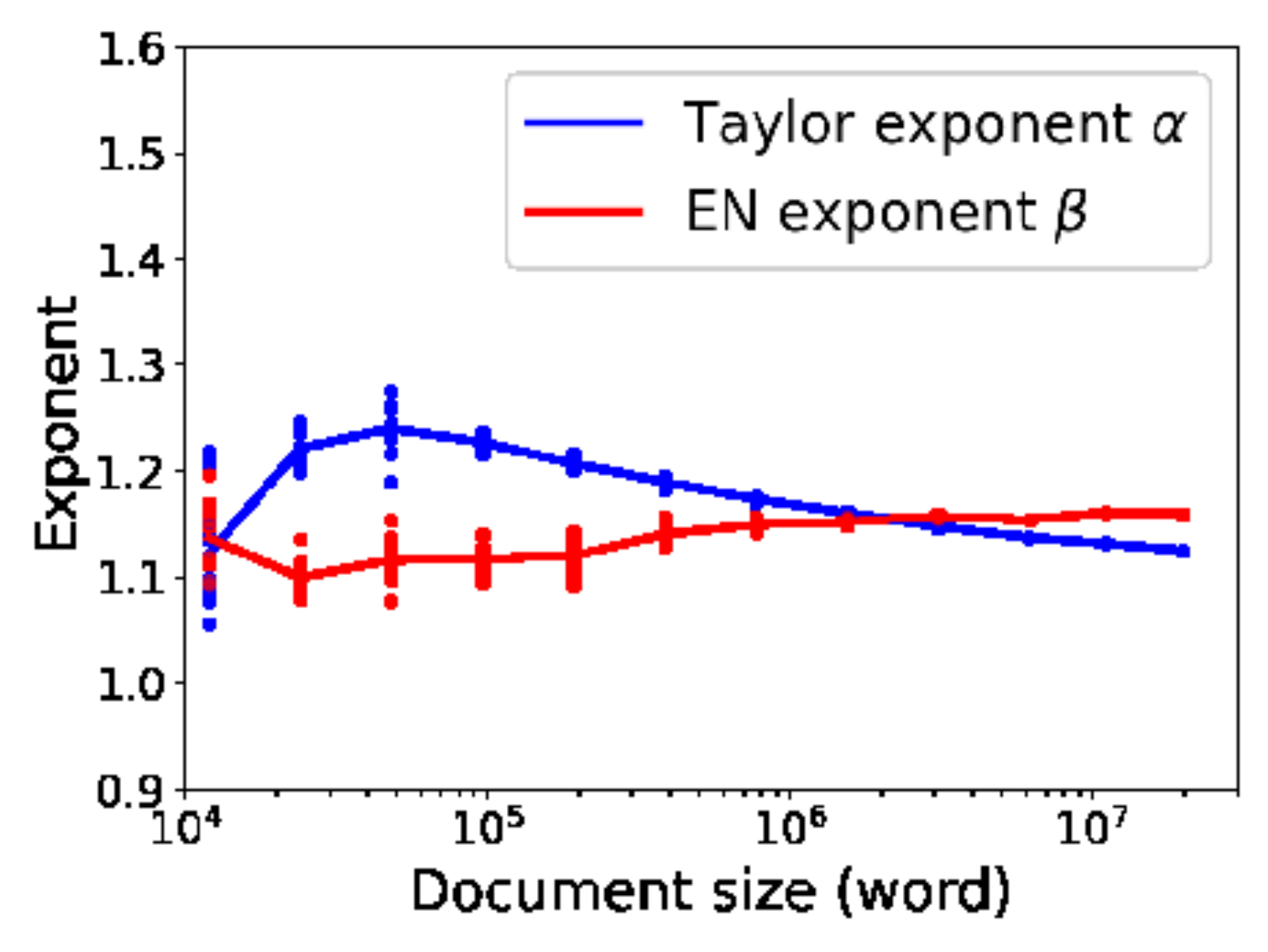}
\caption{\label{fig:docsize} Effect of text size on each method, for
  words in the {\em Wall Street Journal}. For Taylor's method,
  $l=5620$. Ten different portions of the corpus were analyzed for
  each data size except the last three, and the exponents are plotted
  for each portion.}
\end{figure} 

Next, let us examine the influence of the data size on the exponent
value, as shown in \figref{fig:docsize}. The methods were applied at
the word level to a standard large-scale corpus in English, the {\em
  Wall Street Journal} (WSJ), whose statistics are listed in
\tabref{tab:data}. The horizontal axis of \figref{fig:docsize}
indicates the data size in words on a log scale, while the vertical
axis indicates the exponent values of the two methods. The experiment
was conducted on 10 different consecutive portions of the WSJ, except
for the last three points. The last point represents the total size of
the WSJ. Because of this, multiple points are presented vertically,
with the mean behavior represented by the lines.
 
The figure shows that texts must be sufficiently large for a credible
analysis. The plots of both methods are unstable until a length of
$10^5$. This is why we chose the size of the portion of newspaper data
to be $3 \times 10^5$ (see Appendix B) and why the data in
\secref{sec:data} were chosen to be long texts of more than $10^5$
words.
 
Beyond $l=10^5$, the EN exponent appears to stabilize with a slight
increasing tendency. On the other hand, the Taylor exponent slowly
decreases. This decreasing tendency tapers off and seems to converge
as $l$ increases. One reason for this asymptotic behavior lies in the
fact that this graph was produced with a constant $l$. As the document
size increases, $l$ must increase to accommodate the underlying
increase in fluctuation among words.
 
The exponent value was reported to increase with respect to the
logarithm of $l$ \citep{taylor,acl18,jpc18}. Those reports showed how
this tendency applies to any data category, and that the order of the
exponent value remains consistent with respect to text categories. A
similar tendency can be seen in \figref{fig:segment-exponent} in
Appendix C, which plots $\alpha$ versus the segment length $l$ for the
WSJ dataset.

Overall, the tendency of increasing stability for both methods
provides evidence that a dimension underlying fluctuations in a text,
as mentioned later in \secref{sec:fractal}, represents a universal
quality that is almost independent of the size of the text.

\section{Effect of Text Quality} 
The most interesting aspect of the original analysis reported by
Taylor is that he showed how the exponent corresponds to species of
organisms \citep{taylorNature}. Similarly, \citet{acl18,jpc18}
reported that Taylor exponents distinguish text categories. However,
this capability has not been verified for the EN method, because the
original paper only considered a few text samples \citep{Ebeling1995}.

Hence, this section analyzes how the exponents of the two methods
reflect text qualities, in particular the text category and type of
language. The findings indicate that Taylor's method can distinguish
text categories and that the EN method can as well but to a lesser
extent. On the other hand, both methods show some tendency to be able
to capture language types, but not as clearly as text categories.

\subsection{Effect of Text Category} 
\begin{figure} 
\centering \includegraphics[width=\textwidth]{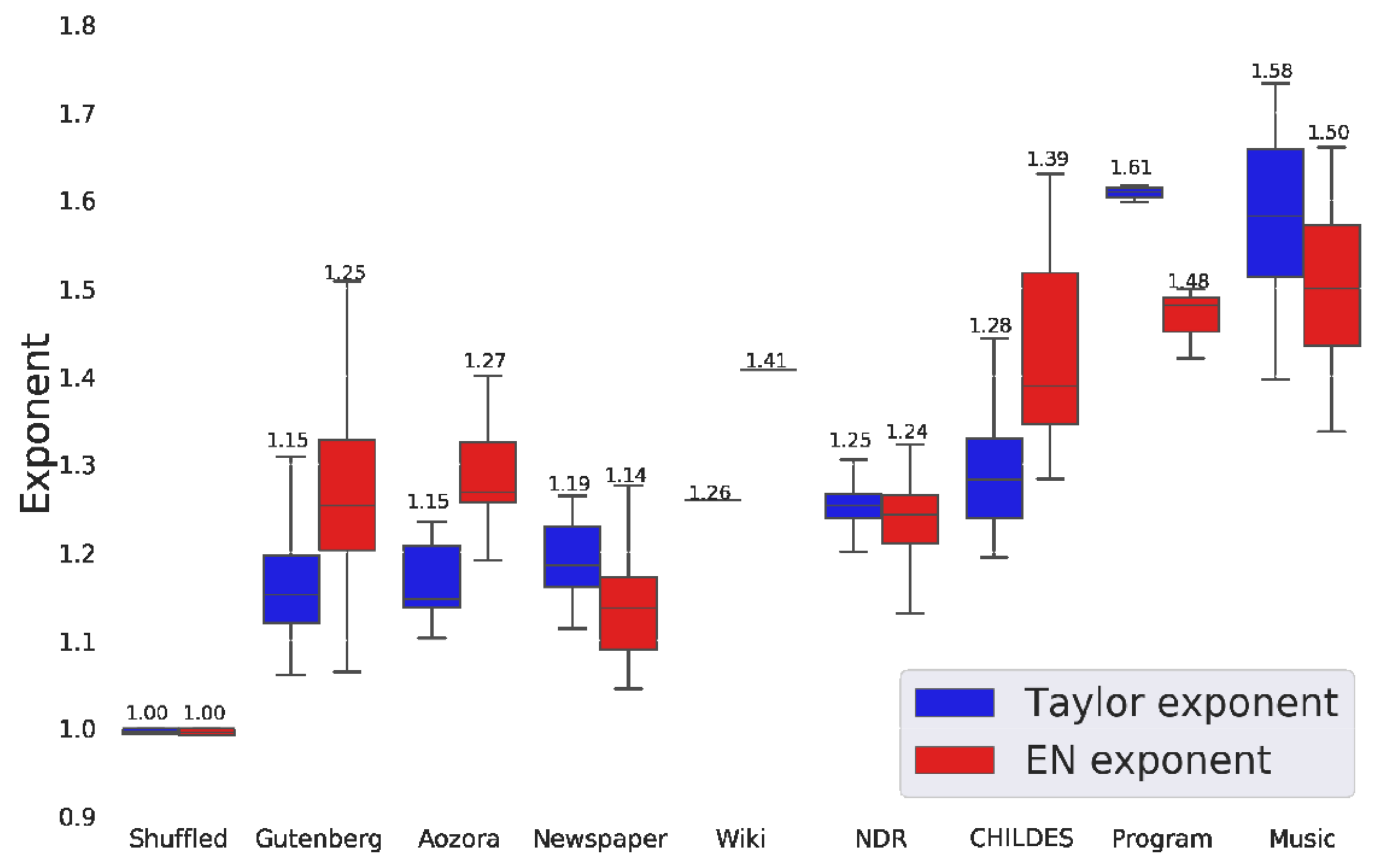}
\caption{\label{fig:box} Boxplots of the exponents of the two methods
  for words in different text categories. }
\end{figure} 

\figref{fig:box} shows the distributions of the exponents with respect
to text categories. The horizontal axis indicates the text categories,
as listed in the first column of \tabref{tab:data}, while the vertical
axis indicates the exponent values. The mean values for every text
category are also listed in \tabref{tab:means} in Appendix D.
 
Each category has a boxplot for each method. The box edges indicate
two quantiles of the data, with the line near the middle of the box
indicating the median. The whiskers indicate the maximum and minimum
values. The number above each maximum whisker is the median value.
 
The two leftmost boxes show the results for shuffled data, consisting
of 10 different samples acquired from {\md}. Following theory, as
explained in \secref{sec:theory}, the two boxes are at a value of
1. The figure thus shows how small the variance of the exponents is in
that case.
 
The rest of the figure shows the results for the real data listed in
\tabref{tab:data}. Most importantly, not a single sample has a value
of 1.
 
Here, we consider the relation between the text category and the
exponents. The second to fourth categories (Gutenberg, Aozora,
Newspaper) are results for written text. The Taylor exponents are
consistently distributed around 1.15, with the newspaper exponents
slightly above those of the literary texts. This is understandable,
because among these written forms, similar phrases appear more
frequently in newspaper texts, and co-occurrences enlarge the exponent
(cf. \secref{sec:theory}). In contrast, the EN exponents vary more
widely. Interestingly, the newspaper exponents are lower for the EN
method, possibly because of the larger number of rare
events. Furthermore, for Project Gutenberg, the variation of the EN
exponents is larger than that of the Taylor exponents.
 
The rest of the figure represents the other categories listed in the
third block of \tabref{tab:data}. The fifth category (Wiki) shows
results for Wikipedia data, which include the Wikipedia
annotations. The Taylor exponents are larger than for the previous
categories, which is a result of grammatical annotations in the form
of wiki tags. Note that those tags have the tendency to increase
co-occurrences (an analysis was given in \secref{sec:theory}). On the
other hand, for the EN method, the increase is very large; the
exponent is 1.41.
 
The sixth category in the figure is speech data (the National Diet
Record, or NDR, in Japanese). The Taylor exponents are larger than for
written text. The EN method shows similar results, but they do not
distinguish the written and spoken text, as the exponent values for
written text range over a large interval.
 
The last three categories consist of language spoken by infants,
music, and programming language. For all three categories, both
exponents are larger than those for the real natural language
texts. Taylor's method gives larger values especially for the
programming language data, because of the large co-occurrence effect
in program source code.
 
We performed a statistical test to evaluate whether the differences
among text categories were significant. Because the number of texts
largely varied among the categories, we conducted a nonparametric
statistical test, the Brunner-Munzel method \citep{bm}\footnote{The
  Kolmogorov-Smirnov test might seem to be a more direct statistical
  test for examining the similarity of distributions, but it presumes
  that a distribution has been captured with many sample points. The
  Brunner-Munzel method is more suitable when the numbers of data
  points differ largely.}. This test examines the difference of the
statistical values acquired from two categories. Two samples, $x_1$
and $x_2$, which are either $\alpha$ or $\beta$ values acquired from
two sample texts, are each taken from the categories under comparison;
then, the null hypothesis is that the probability of $x_1 > x_2$ and
that of $x_2 > x_1$ are equal.

All categories with at least 10 samples were tested (the Wikipedia and
program categories were not tested). Appendix E lists the p-values
(\tabref{test-p-taylor-category} for Taylor's method, and
\tabref{test-p-ebeling-category} for the EN method). The results
conform to our expectations with a significance level of 0.05. For
Taylor's method, the null hypothesis was validated for the written
text (Gutenberg-Aozora, Aozora-Newspaper) and speech (CHILDES-NDR)
data \footnote{The null hypothesis was rejected for the Gutenberg-News
  samples, possibly because these datasets have very large differences
  in the number of samples.}. The null hypothesis was rejected for all
other pairs. On the other hand, for the EN method, the null hypothesis
was rejected for all pairs except Gutenberg-Aozora. This suggests that
the EN method could not capture category differences.

Overall, we can say that Taylor's method captures categories much
better than the EN method does. Such a characteristic captured across
categories seems to correspond to some degree of co-occurrence in
texts. The correspondence from written to spoken and to other
categories suggests a change in complexity. The results presented in
this section show how that change is captured by the Taylor exponent
but not by the EN exponent. However, the rough results of the EN
method are consistent with those of Taylor's method, giving larger
values for the CHILDES, program, and music data than for written
texts.

\subsection{Effect of Type of Language} 
\label{sec:language} 
Because our data consisted of various language types, we also examined
the exponents in relation to language type.

\begin{figure}[t] 
	\begin{subfigure}[t]{.49\linewidth} 
	\centering
        \includegraphics[width=80mm]{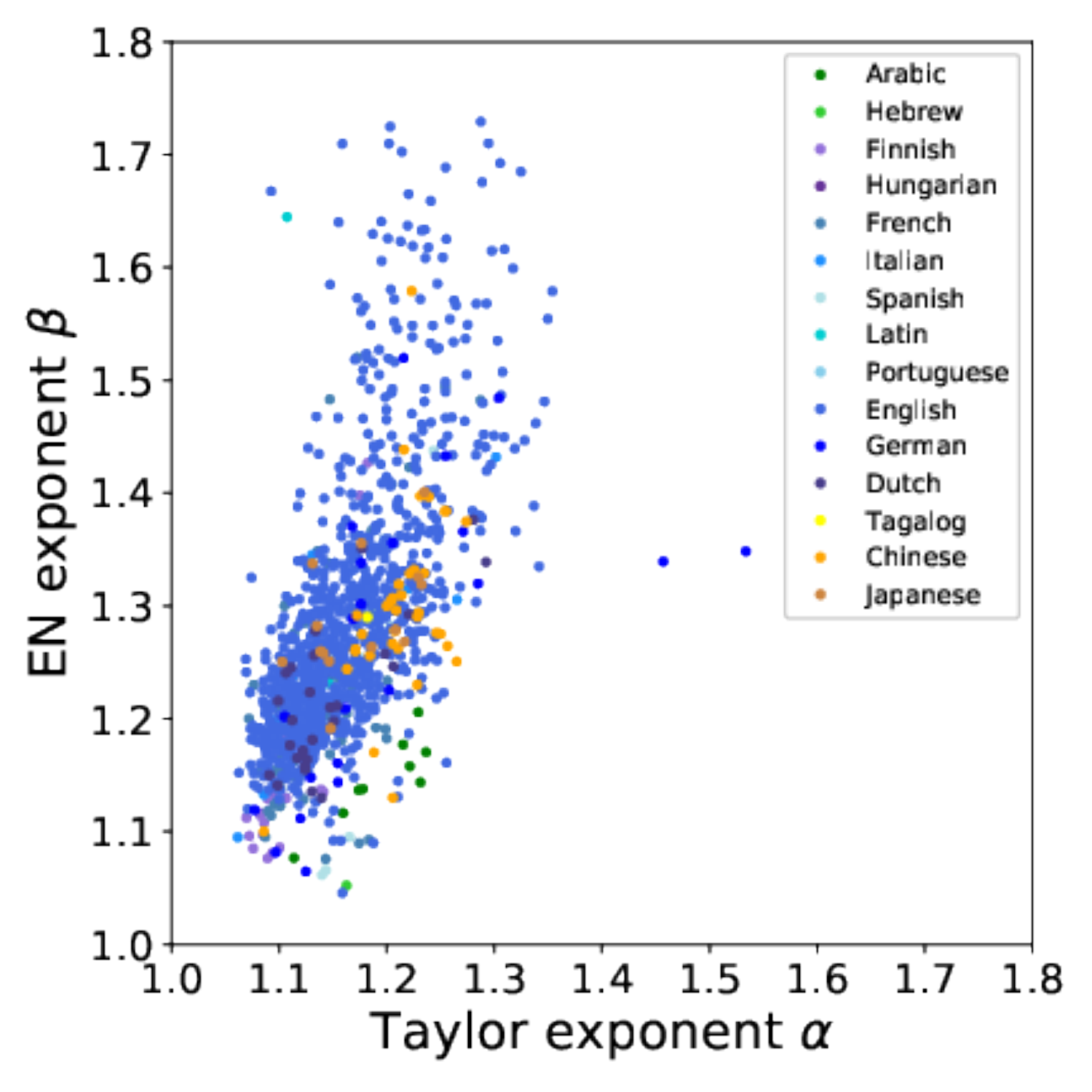}
	\caption{Word level}
	\end{subfigure}  
	\begin{subfigure}[t]{.49\linewidth} 
	\centering
        \includegraphics[width=80mm]{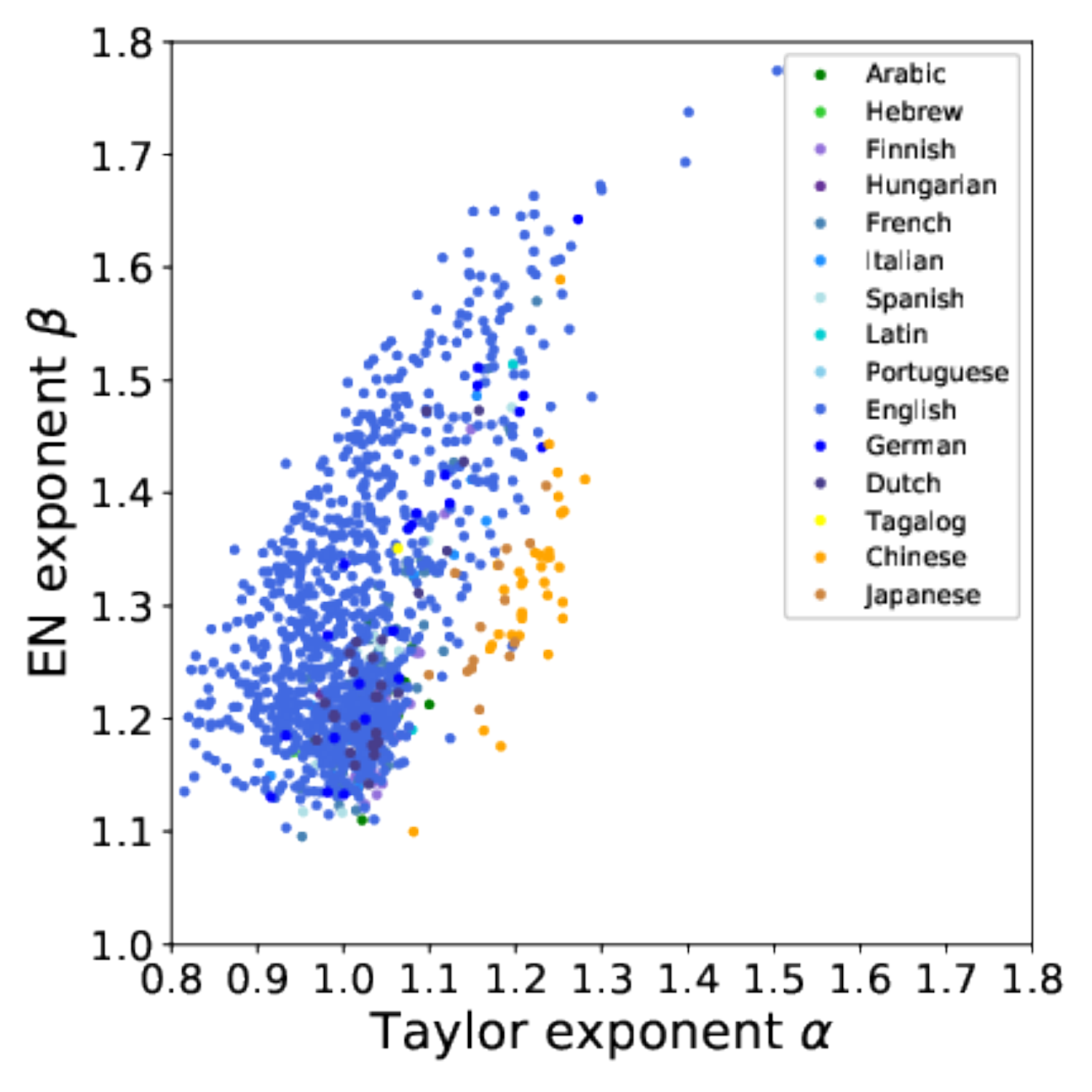}
	\caption{Character level}
	\end{subfigure} 
	\caption{\label{fig:scatter-lang}Scatter plot of EN and Taylor
          exponents for different languages.}
\end{figure} 

\label{sec:characters} 

\figref{fig:scatter-lang} shows scatter plots of the Taylor and EN
exponents for all of the natural-language written texts listed in the
first and second blocks of \tabref{tab:data}. Each point corresponds
to a text, and the Taylor exponent $\alpha$ and EN exponent $\beta$
are indicated by the horizontal and vertical coordinates,
respectively. The color of each point represents the text's
language. Because the majority of texts are in English, the
English texts are in the background, whereas those of the other
languages are in the foreground.
 
The left graph shows the results for words. It indicates the extent to
which the EN and Taylor exponents correlate. The Spearman correlation
of the two methods is 0.69; thus, the exponents are strongly
correlated. The graph also shows how the EN method covers a wider
range of exponents compared with Taylor's method (\figref{fig:box}
also showed this).
 
The graph seems to show a rough clustering of languages, such as the
orange points for Chinese and Japanese at the middle right. Therefore,
both exponents show some tendency to distinguish languages.

On the other hand, the right graph shows the results for
characters. The Spearman correlation between $\alpha$ and $\beta$ is
0.44, lower than that for words but still indicating a significant
correlation.

The Taylor exponent is around 1 for many of the texts in alphabetic
scripts. The analytical results presented in \secref{sec:analysis}
indicate that Taylor's method cannot distinguish texts in alphabetic
scripts from an i.i.d. sequence. This is natural, because alphabetic
scripts often have fewer than 100 characters, which is too few for a
Taylor analysis of the distribution of character types. Once the
number of characters reaches the level of Chinese and Japanese,
though, the exponents are clearly larger than 1, at around 1.2, which
is close to the exponent values of the written texts.
 
As for the EN method, all the natural language texts have exponents
above 1. The EN method thus distinguishes real texts from
i.i.d. sequences at the character level. Furthermore, the graph shows
that the blue points are clustered at the bottom (around 1.15 on the
vertical scale).
 
\begin{figure}[t] 
	\centering
        \includegraphics[width=120mm]{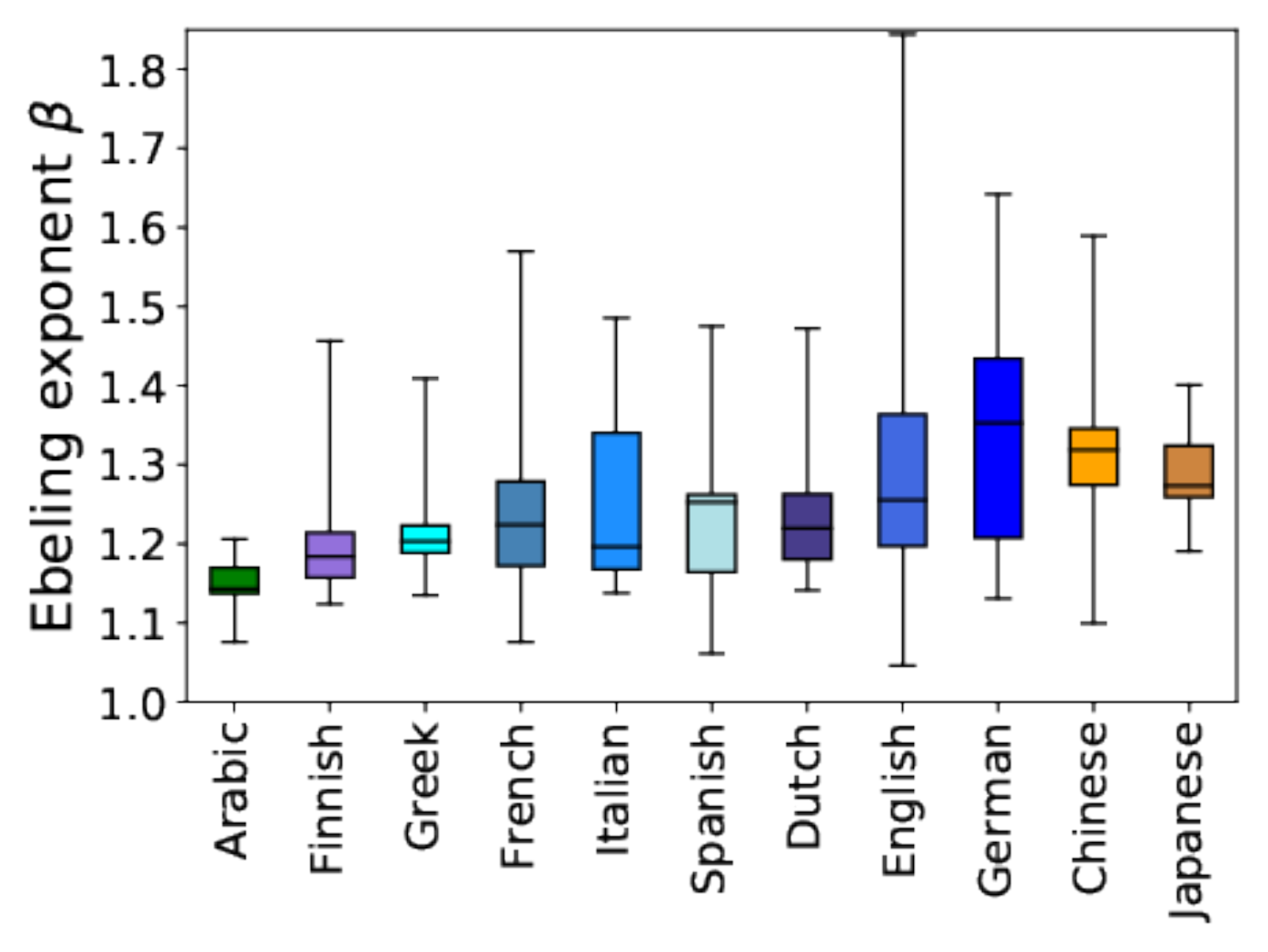}
	\caption{EN exponents for characters in different language
          scripts.}
        \label{fig:enlanguage} 
\end{figure} 
 
\figref{fig:enlanguage} shows boxplots of texts for characters with
respect to language for the EN exponent $\beta$. Only results for
languages with more than 10 samples are included. From left to right,
the languages range from consonantal (Arabic, Hebrew) to alphabetic
scripts and ideographs (Chinese and Japanese). The languages are not
easy to distinguish; the ranges for English and German include those
for Chinese and Japanese.
 
The statistical test by the Brunner-Munzel method reflects this
reality. Here, too, only languages with more than 10 samples were
included, and each pair of languages was tested as to whether the
exponents were the same. It would be interesting if the test could
capture differences among different scripts, but Appendix E shows that
the results are not clear. For words, the results for many random
language pairs appear to validate the null hypothesis. For characters,
the results of the EN method are slightly better: for example, the
Japanese-Chinese and Spanish-Italian pairs validate the null
hypothesis, but the test results appear almost random.

\section{Discussion} 
\subsection{Empirical Pros and Cons of the Two Methods} 
Thus far, we have discussed the differences in the EN and Taylor
methods in terms of qualitative and quantitative evidence. Moreover,
Taylor's method corresponds to the first phase of the EN method, and
hence, the methods share a general tendency. Here, we will highlight
their differences.

The EN method is statistically more stable. The exponent remains
relatively invariant across data sizes. In particular, the method can
distinguish a text from an i.i.d. process at any element level. This
quality results from summing the variances of all element types. For
the same reason, the EN exponent can capture the text category only
very roughly.

In contrast, Taylor's method can distinguish natural language text
from an i.i.d. process only at the word level. On the other hand, at
the character level, it is not applicable to a target system with a
small number of elements, such as an alphabet or a consonantal script,
because it considers the distribution of elements. It can, however,
distinguish text categories when it is applied to words.

Overall, although Taylor's method shows some dependency on parameters,
if the set of elements is large enough, it is a more direct analysis
than the EN method. It also has the possibility of highlighting
qualitative differences as categorical distinctions. If the element
set consists of fewer than 100 elements, then the EN method can reveal
some properties of the system, although the exponent only reflects
rough qualitative characteristics of a text.

It is important to learn what characteristics a method highlights. In
this study, we revealed subtle differences between the results of
Taylor's method and the EN method when they are used on natural
language text, a good target for this purpose as it is interpretable
in various ways. We believe that differences such as the ones found
here would appear when the methods are applied to targets other than
language.

\subsection{Interpretation of Exponents as Fractal Dimension} 
\label{sec:fractal} 
The resulting exponents can be interpreted as a {\em dimension}
underlying the fluctuation of language. The EN exponent especially has
an affinity with the similarity dimension, as a kind of fractal
dimension \citep{fractal1,fractal2}. The similarity dimension is
defined for a geometrical object in a metric space. If we suppose that
enlarging an object $a$ times is equivalent to making $b$ copies of
the original object, the similarity dimension can be defined as
follows:
\begin{equation} 
  D = \log b / \log a.
\end{equation} 
For example, for a square whose edge length is $m$, doubling $m$ would
give a square $a=2$ times larger by edge length, but filling this
larger square requires $b=4$ of the original squares. Therefore, the
similarity dimension is $D=\log 4 / \log 2$ = 2. Similarly, the
similarity dimension of a cube is $D=3$. A famous example is the Koch
curve, whose similarity dimension is $D = \log 4 / \log 3 =
1.262$. For a complex object, this dimension $D$ is known not to be an
integer.

A text is not a geometrical object, so its length cannot be treated in
the same way as a geometrical object defined in a metric
space. Nevertheless, it is also true that we often say that a text has
a {\em length}. The Taylor and EN exponents then suggest a
metaphorical interpretation with respect to the fractal dimension. The
EN case shows that a text portion that is $m$ times longer has
$m^\beta$ times more fluctuation. Accordingly, $D = \log m^\beta /
\log m = \beta$. Similarly, although the Taylor exponent cannot be
interpreted as the similarity dimension, it shows how words occurring
$m$ times more frequently would fluctuate $m^\alpha$ times more.

The Taylor and EN analyses we have presented here imply that a text
does not simply consist of its portions. When the portions are
concatenated, they require further editing to unify them into a
whole. That is, a text requires linkages across its portions. These
linkages generate holistic fluctuations, causing $\alpha$ and $\beta$
not to be 1. A text therefore has a property of amplifying
fluctuations in a self-similar manner, and Taylor and EN analyses are
capable of revealing this property.

\section{Conclusion} 
This article considered the commonalities and differences between
Taylor's method and the Ebeling \& Neiman (EN) method in a large-scale
study of natural language texts and related data.

Taylor's method analyzes the distribution of the variance of every
element type with respect to the mean within a given segment
length. On the other hand, the EN method analyzes how the sum of the
variances depends on the segment length. The results of these methods
are power laws having respectively Taylor or EN exponents. Either
exponent is 1 for an i.i.d. sequence. On the other hand, when a
sequence presents some clustering tendency, the exponents become
larger: in some cases, such as with consistent co-occurrences and
clustered rare words, the exponents reach 2.

Because Taylor's method can be regarded as the first step of the EN
method, the outcomes of the two methods are correlated. Nevertheless,
there are differences that derive from the methods'
procedures. Because Taylor's method is a direct analysis of the
distribution of the variances among elements, it can distinguish text
categories when it is applied to words. Such a distinction is not so
clear with the EN method, possibly because it accumulates the
variances of all element types. Both methods show some limited
possibility of being able to distinguish languages.

Our findings are based on natural language text as the target of
analysis; we chose text because it is interpretable from various
aspects. Moreover, we believe that the findings presented in this
article apply to other analysis targets besides natural language.

\bibliographystyle{natbib} \bibliography{en}

\section*{Appendix A: Procedure to measure mean and variance} 
Let us segment a sequence $X$ into $\lfloor{\frac{N}{l}}\rfloor$
segments of length $l$ with no overlap. The appearances of an element
$w$ are counted in each segment, and the mean and variance are
computed from the counts.

$l$ is chosen as follows. For Taylor's method, $l$ must be
sufficiently smaller than the document length in order to calculate
the variance; in other respects, its choice is arbitrary. Among the
different values of $l$ taken from logarithmic bins, we chose the
maximum $l$ that could apply to all documents, specifically $l = 5620
\approx 10^{3.75}$. For the EN method, the points were taken from
segments with an exponentially growing length, e.g., $l_j=K a^{j-1}$,
$j=1,\ldots,J$, where $K \leq l_j < L$. We set $L \equiv
\min(10000,N)$, $a=1.8$, $K=10$.

\section*{Appendix B: Datasets} 
\label{sec:appB} 

\begin{table}[ht]
  \caption{Lengths and mean vocabulary sizes of texts.}
  \label{tab:stat}
  \label{tab:data}
  \footnotesize
\centering
\begin{tabular}{|p{3cm}|c|c|r|r|r|r|}
\hline
\multicolumn{1}{|c|}{\multirow{2}{*}{Texts   (Category)}} & \multirow{2}{*}{Language} & \#                           & \multicolumn{2}{c|}{Total Length}                     & \multicolumn{2}{c|}{Mean Vocab Size}                  \\ \cline{4-7} 
\multicolumn{1}{|c|}{}                                    &                           & \multicolumn{1}{r|}{samples} & \multicolumn{1}{c|}{word} & \multicolumn{1}{c|}{char} & \multicolumn{1}{c|}{word} & \multicolumn{1}{c|}{char} \\ \hline
                                                          & English                   & 910                          & 284945934                 & 1421317443                & 17237.7                   & 87.4                      \\ \cline{2-7} 
                                                          & French                    & 66                           & 19350770                  & 102588196                 & 22098.3                   & 105.0                     \\ \cline{2-7} 
                                                          & Finnish                   & 33                           & 6518170                   & 42270908                  & 33597.1                   & 86.6                      \\ \cline{2-7} 
                                                          & Chinese                   & 32                           & 20157338                  & 43006836                  & 15352.9                   & 4413.0                    \\ \cline{2-7} 
                                                          & Dutch                     & 27                           & 6935199                   & 37021318                  & 19159.1                   & 97.1                      \\ \cline{2-7} 
                                                          & German                    & 20                           & 4723500                   & 26590388                  & 24242.3                   & 115.2                     \\ \cline{2-7} 
Gutenberg (written)                                       & Italian                   & 14                           & 3735326                   & 19990753                  & 29103.5                   & 101.7                     \\ \cline{2-7} 
                                                          & Spanish                   & 12                           & 4366047                   & 21653921                  & 26111.1                   & 101.3                     \\ \cline{2-7} 
                                                          & Greek                     & 10                           & 1599692                   & 8963014                   & 22805.7                   & 142.7                     \\ \cline{2-7} 
                                                          & Latin                     & 2                            & 1011487                   & 4868576                   & 59667.5                   & 282.0                     \\ \cline{2-7} 
                                                          & Portuguese                & 1                            & 261382                    & 1333023.0                 & 24719.0                   & 110.0                     \\ \cline{2-7} 
                                                          & Hungarian                 & 1                            & 198303                    & 1037517.0                 & 38384.0                   & 104.0                     \\ \cline{2-7} 
                                                          & Tagalog                   & 1                            & 208455                    & 1193099.0                 & 26335.0                   & 109.0                     \\ \hline
Moby Dick                                                 & English                   & 1                            & 254655                    & 1255837                   & 20473.0                   & 78.0                      \\ \hline
Aozora (written)                                          & Japanese                  & 13                           & 8016804                   & 20349717                  & 19760.0                   & 3050.5                    \\ \hline
\hline
                                                          & Arabic                    & 90                           & 27008988                  & 129553552                 & 25200.9                   & 93.4                      \\ \cline{2-7} 
                                                          & English                   & 70                           & 21006597                  & 115764317                 & 21407.3                   & 80.1                      \\ \cline{2-7} 
Newspaper (written)                                       & Chinese                   & 30                           & 9003127                   & 22558321                  & 18850.3                   & 3494.1                    \\ \cline{2-7} 
                                                          & French                    & 30                           & 9002942                   & 50625194                  & 28134.9                   & 99.1                      \\ \cline{2-7} 
                                                          & Hebrew                    & 10                           & 3001046                   & 14998734                  & 46560.6                   & 50.0                      \\ \cline{2-7} 
                                                          & Japanese                  & 10                           & 3001150                   & 7737532                   & 19833.0                   & 2474.0                    \\ \hline
Wall Street Journal                                       & English                   & 1                            & 22679513                  & 117305668                 & 137467.0                  & 87.0                      \\ \hline
\hline
Wiki (enwiki8)                                                   & tag-annotated             & 1                            & 14647848                  & -                         & 1430791.0                 & -                         \\ \hline
NDR (National                                             & Japanese                  & 392                          & 121781033                 & 201204976                 & 8882.9                    & 1964.0                    \\
Diet Record, speech)                                      &                           &                              &                           &                           &                           &                           \\ \hline
CHILDES (speech)                                          & various                   & 10                           & 1934340                   & -                         & 9908.0                    & -                         \\ \hline
Programs                                                  & various                   & 4                            & 136644076                 & -                         & 838907.8                  & -                         \\ \hline
Music                                                     & MIDI                      & 12                           & 1631921                   & -                         & 9187.9                    & -                         \\ \hline
\end{tabular}\end{table}

The datasets are publicly available, as listed in
\tabref{tab:stat}. In general, the data of Taylor or EN analyses
should be chosen carefully, because these analyses require a large
amount of data, as illustrated in \figref{fig:docsize}. A large-scale
text often includes repetitions of large chunks. For such bad data,
the analysis would only indicate the poor quality of the data rather
than the true nature of the method. From this perspective, the data
used in this article were chosen under strict criteria.

The data were categorized into the following three types:
single-author literary texts (first block in \tabref{tab:stat}),
multi-author newspaper corpora (second block), and other kinds of data
related to language (third block). The columns of \tabref{tab:stat}
indicate the kind of data, the kind of language, the number of
samples, the total length (words and characters), and the mean
vocabulary size (words and characters). For example, 910 English
literary texts were extracted from Project Gutenberg, and the {\em
  Total Length} column lists the {\em total length} in words for all
910 texts. Those texts have on average 313,127 words (= 284,945,934 /
910) and 1,42,317,443 characters (= 1,421,317,443 / 910 = 1,561,887),
and mean vocabulary sizes of 17,238 words and 87 characters.

The first block of \tabref{tab:data} lists the names of the literary
texts extracted from Project Gutenberg and Aozora Bunko. The row for
{\em Moby Dick} after Project Gutenberg is the example in the main
text (\secref{sec:example}). The set of literary texts consists of
single-author texts extracted from the Project Gutenberg and Aozora
Bunko archives and covers 14 languages\footnote{Project Gutenberg
  includes hardly any Japanese literary texts; the Japanese texts were
  thus acquired from Aozora Bunko.}. We initially collected texts
above a certain size threshold (1 MB, including annotations). Project
Gutenberg splits many of the texts into different volumes; these were
manually combined into single texts.

Once the texts were collected, their metadata annotations were
eliminated. The
PyNLPIR \footnote{\tt{https://github.com/jordwest/mecab-docs-en}} and
MeCab \footnote{\tt{https://pypi.org/project/PyNLPIR/}} applications
were used to segment the Chinese and Japanese texts for the word-based
analyses. NLTK\footnote{{\tt https://www.nltk.org}} was used to
tokenize the texts in the other languages.

The word set is often difficult to define: accurate lemmatization is
not available for some non-major languages, and it is difficult to
capture all the customs of writing systems across languages. We
considered that the fairest cross-language approach was to use all
tokenized results, without introducing any arbitrary elimination
scheme. In other words, conjugated words, capitalized words,
misspelled words, and symbols at the word level were all included in
the sequences used to produce the results in this article.

\begin{table}[] 
\caption{List of newspaper corpora used in the experiments.}
\label{tab:news} 
\centering
\begin{tabular}{|l|p{10cm}|} 
\hline Language & \multicolumn{1}{c|}{Newspaper} \\ \hline English &
Wall Street Journal, New York Times, Central News Agency of Taiwan
(English), Associated Press (English), Xinhua News Agency (English),
Agence-France Presse (English), Los Angeles Times, Washington Post,
Washington Post/Bloomberg \\ \hline Arabic & Daily Aaj, Agence
France-Presse (Arabic), Al-Ahram, Assabah, Asharq Al-Awsat, Al Hayat,
Al-Quds Al-Arabi, Ummah Press, Xinhua News Agency (Arabic) \\ \hline
Chinese & People's Daily, Xinhua News Agency, Central News Agency of
Taiwan\\ \hline French & Le Monde, Associated Press (French), Agence
France-Presse (French) \\ \hline Hebrew & Haaretz \\ \hline Japanese &
Mainichi \\ \hline
\end{tabular} 
\end{table} 

The second block of \tabref{tab:data} lists data from newspapers,
which are characterized as multi-author texts. The above preprocessing
scheme was applied to all of the natural language data from these
newspapers. \tabref{tab:news} lists the newspapers used. Many of them
are distributed from credible sources such as the Language Data
Consortium. In the case of the {\em Wall Street Journal} in English,
all of the text was used, because it is a standard corpus of highly
credible quality. For the other newspapers, 10 non-overlapping
portions were extracted in units of articles from random parts of the
corpus, so that each portion contained at least 300,000 words. This
value of 300,000 words was chosen according to the statistical
analysis described in \secref{sec:basic_properties}, and it met the
criteria discussed there.

\cite{daniels} defined six kinds of writing system used around the
globe: alphabetic (Indo-European scripts), abjad (Arabic, Hebrew),
abugida (Indian and south Asian scripts), logosyllabary (Chinese),
syllabary (Japanese kana), and featural (Korean). This article does
not consider the block characters of an abugida or a featural script,
because in those cases, one block character is constructed from parts,
and the notion of what is ``one character'' is deemed ambiguous, as a
character can be either a component or a combination.

The third block lists the other language-related samples considered in
this article, which were introduced for the purpose of comparing the
behaviors of the exponents of the two methods.

In the third block, the first row indicates the enwiki8 100-MB dump
datasets of Wikipedia, represented by the tag Wiki, consisting of
tag-annotated text from English Wikipedia. These data {\em included}
the annotations made by Wikipedia, for the purpose of observing their
effect.
 
The second row indicates speech data. Obtaining long, clean data is an
issue for speech, because speech content changes over time, so
sessions are often very short. From the results of a search through
various data sources, this article includes data of the National Diet
Record (NDR) in Japanese. Records of 250 sessions were used, with each
session corresponding to an opening of a National Diet meeting. These
data are long and clean; they originally consisted of speech data that
were transcribed by professionals.

The third row in the block lists data of the Child Language Data
Exchange System (CHILDES). The 10 longest child-directed speech
utterances in the CHILDES database were used\footnote{The 10 longest
  utterances were as follows: Thomas \citep{childes,childes-english},
  Groningen \citep{childes-dutch}, Rondal \citep{childes-french}, Leo
  \citep{childes-german}, Ris \citep{childes-indonesian}, Nanami
  \citep{childes-japanese}, Inka \citep{childes-polish}, Angela
  \citep{childes-serbian}, Beca \citep{childes-spanish}, and Boteborg
  \citep{childes-swedish}.}. The data were preprocessed by extracting
only the children's utterances.

The fourth row lists program source-code data (in the Lisp, Haskell,
C++, and Python programming languages) crawled from large
representative archives. The data were parsed and stripped of comments
written in natural language. The data contain many repetitions,
because of the general tendency of programmers to copy and paste
sample code. In this sense, the results in this article should be
considered only one possible result. It remains an open question how
best to reuse programming source code as text data.

Finally, the last row in the third block represents 12 pieces of
musical data (long symphonies and so forth) that were transformed from
MIDI data into text with the SMF2MML
software \footnote{\url{http://shaw.la.coocan.jp/smf2mml/}}; the
annotations were then removed. SMF2MML uses a descriptive method to
transcribe the set of sounds occurring at a point in time together
with their duration into a word. The transcription approach in this
article is only one of the many possible for music.

\section*{Appendix C: Dependency of Taylor Exponent on $l$} 
\label{sec:taylorL} 
As reported in \cite{taylor}, the Taylor exponent $\alpha$ grows as
the segment length $l$ increases. For texts, it was reported in
\cite{acl18,jpc18} that $\alpha$ depends on the logarithm of $l$. This
appendix provides evidence of this from the {\em Wall Street Journal}
(the data are listed in the first row of the second block in
\tabref{tab:data}).
 
\begin{figure}[h] 
\centering \includegraphics[width=80mm]{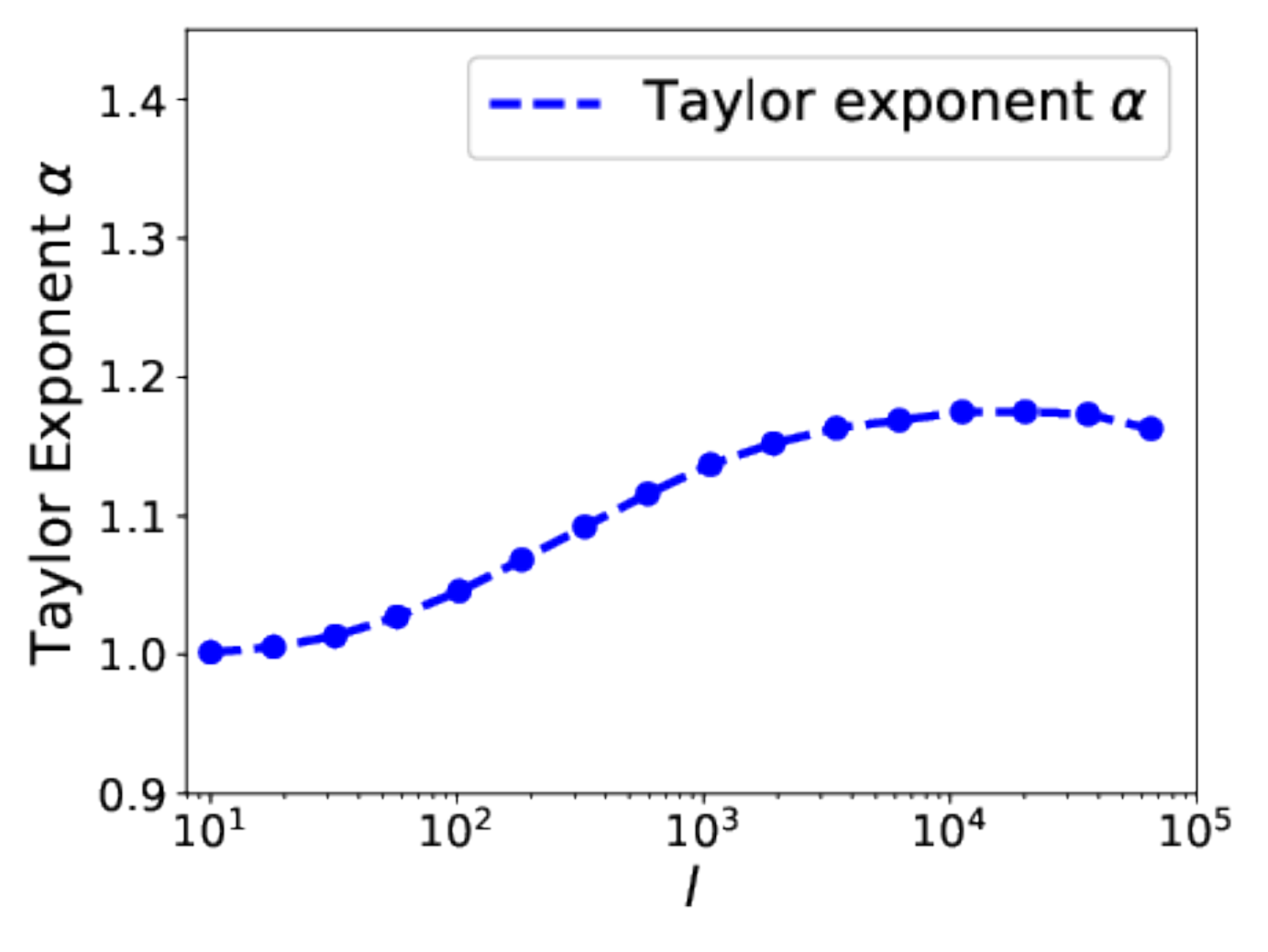}
\caption{\label{fig:segment-exponent} Taylor exponent with respect to
  the segment length $l$.}
\end{figure} 
 
\figref{fig:segment-exponent} shows the Taylor exponent with respect
to $l$. The horizontal axis indicates the logarithm of the segment
length, while the vertical axis indicates the Taylor exponent. The
figure shows an increasing tendency with respect to $l$, with
saturation occurring around $10^4$. This is due to the fact that a
large $l$ close to $10^5$ causes the number of segments to
decrease. The Taylor exponent is initially $1$ when the segment length
is $1$. This can be analytically explained as follows
\citep{taylor}. Consider the case of $l=1$. Let $n$ be the frequency
of a particular word in a segment. We have $\average{n} \ll 1$,
because the probability of a specific word appearing in a segment
becomes very small. Because $\average{n}^2 \approx 0$, $\sigma^2 =
\average{n^2} - \average{n}^2 \approx \average{n^2}$. Then, because
$n=1$ or $0$ (with $l$=1), $\average{n^2} = \average{n} = \mu$. Thus,
$\sigma^{2} \approx \mu$.

As mentioned in the main text, Taylor's method can be regarded as the
first step of the EN method. \figref{fig:segment-exponent} shows that
the Taylor exponent w.r.t. $l$ corresponds to $z(l)$ w.r.t $l$.

\newpage 

\section*{Appendix D: Mean Taylor and EN Exponents} 
\label{sec:meanvalues} 

\begin{table}[h]
  \caption{Mean exponent values of texts for each category listed in \tabref{tab:data}.
    \label{tab:means}}
  
\centering
\begin{tabular}{|l|c|c|l|c|l|}
\hline
\multicolumn{1}{|c|}{\multirow{2}{*}{Texts}} & \multirow{2}{*}{Language} & \multicolumn{2}{c|}{$\alpha$} & \multicolumn{2}{c|}{$\beta$}  \\ \cline{3-6} 
\multicolumn{1}{|c|}{}                       &                           & word & char                   & word & char                   \\ \hline
                                             & English                   & 1.16 & 1.01                   & 1.29 & 1.29                   \\ \cline{2-6} 
                                             & French                    & 1.14 & 1.04                   & 1.23 & 1.24                   \\ \cline{2-6} 
                                             & Finnish                   & 1.10 & 1.04                   & 1.17 & 1.20                   \\ \cline{2-6} 
                                             & Chinese                   & 1.22 & 1.22                   & 1.31 & 1.33                   \\ \cline{2-6} 
                                             & Dutch                     & 1.14 & 1.04                   & 1.22 & 1.24                   \\ \cline{2-6} 
                                             & German                    & 1.18 & 1.07                   & 1.27 & 1.33                   \\ \cline{2-6} 
Gutenberg                                    & Italian                   & 1.14 & 1.05                   & 1.24 & 1.25                   \\ \cline{2-6} 
                                             & Spanish                   & 1.16 & 1.04                   & 1.25 & 1.26                   \\ \cline{2-6} 
                                             & Greek                     & 1.16 & 1.02                   & 1.18 & 1.22                   \\ \cline{2-6} 
                                             & Latin                     & 1.14 & 1.14                   & 1.44 & 1.35                   \\ \cline{2-6} 
                                             & Portuguese                & 1.12 & 1.06                   & 1.20 & 1.35                   \\ \cline{2-6} 
                                             & Hungarian                 & 1.14 & 0.99                   & 1.24 & 1.22                   \\ \cline{2-6} 
                                             & Tagalog                   & 1.18 & 1.06                   & 1.29 & 1.35                   \\ \hline
Moby Dick                                    & English                   & 1.15 & 1.03                   & 1.25 & 1.22                   \\ \hline
Aozora                                       & Japanese                  & 1.18 & 1.17                   & 1.29 & 1.29                   \\ \hline
                                             & Arabic                    & 1.19 & 1.04                   & 1.18 & 1.21                   \\ \cline{2-6} 
                                             & English                   & 1.18 & 0.98                   & 1.14 & 1.12                   \\ \cline{2-6} 
Newspaper                                    & Chinese                   & 1.22 & 1.22                   & 1.17 & 1.19                   \\ \cline{2-6} 
                                             & French                    & 1.13 & 1.00                   & 1.15 & 1.13                   \\ \cline{2-6} 
                                             & Hebrew                    & 1.15 & 0.97                   & 1.09 & 1.13                   \\ \cline{2-6} 
                                             & Japanese                  & 1.24 & 1.18                   & 1.26 & 1.30                   \\ \hline
National Diet Record                         & Japanese                  & 1.25 & 1.22                   & 1.21 & 1.23                   \\ \hline
enwiki8                                      & tag-annotated             & 1.26 & \multicolumn{1}{c|}{-} & 1.41 & \multicolumn{1}{c|}{-} \\ \hline
CHILDES                                      & various                   & 1.36 & \multicolumn{1}{c|}{-} & 1.43 & \multicolumn{1}{c|}{-} \\ \hline
Programs                                     & various                   & 1.58 & \multicolumn{1}{c|}{-} & 1.47 & \multicolumn{1}{c|}{-} \\ \hline
Music                                        & MIDI                      & 1.58 & \multicolumn{1}{c|}{-} & 1.51 & \multicolumn{1}{c|}{-} \\ \hline
\end{tabular}
\end{table}

\tabref{tab:means} lists the mean Taylor exponent $\alpha$ and EN
exponent $\beta$ for all the data categories listed in
\tabref{tab:data}.
 
\newpage 
 
\section*{Appendix E: Statistical Test Results for Text Category Pairs} 
\label{sec:quanlltiles} 
\tabref{test-p-taylor-category} and \tabref{test-p-ebeling-category}
list the p-values from the statistical test of the Brunner-Munzel
method for every pair of categories with at least 10 samples.

\begin{table}[h]
\centering
\caption{p-values of the Brunner-Manzel test for the Taylor exponent and category (Wiki, Music, and Program categories omitted because of small sample sizes).}
\label{test-p-taylor-category}
\begin{tabular}{|l|c|c|c|c|c|c|c|}
\hline
          & \multicolumn{1}{l|}{Shuffled} & \multicolumn{1}{l|}{Gutenberg} & \multicolumn{1}{l|}{Aozora} & \multicolumn{1}{l|}{News} & \multicolumn{1}{l|}{NDR} & \multicolumn{1}{l|}{CHILDES} & \multicolumn{1}{l|}{Music} \\ \hline
Shuffled  & N/A                           & 0.000                          & 0.000                       & 0.000                     & 0.000                    & 0.000                        & 0.000                      \\ \hline
Gutenberg & -                             & N/A                            & 0.347                       & 0.000                     & 0.000                    & 0.000                        & 0.000                      \\ \hline
Aozora    & -                             & -                              & N/A                         & 0.205                     & 0.000                    & 0.000                        & 0.000                      \\ \hline
Newspaper & -                             & -                              & -                           & N/A                       & 0.000                    & 0.000                        & 0.000                      \\ \hline
NDR       & -                             & -                              & -                           & -                         & N/A                      & 0.116                        & 0.000                      \\ \hline
CHILDES   & -                             & -                              & -                           & -                         & -                        & N/A                          & 0.000                      \\ \hline
Music     & -                             & -                              & -                           & -                         & -                        & -                            & N/A                        \\ \hline
\end{tabular}
\end{table}

\begin{table}[h]
\centering
\caption{p-values of the Brunner-Manzel test for the EN exponent and category (Wiki, Music, and Program categories omitted because of small sample sizes).}  
\label{test-p-ebeling-category}
\begin{tabular}{|c|c|c|c|c|c|c|c|}
\hline
          & Shuffled & Gutenberg & Aozora & News  & NDR   & CHILDES & Music \\ \hline
Shuffled  & N/A      & 0.000     & 0.000  & 0.000 & 0.000 & 0.000   & 0.000 \\ \hline
Gutenberg & -        & N/A       & 0.084  & 0.000 & 0.000 & 0.000   & 0.000 \\ \hline
Aozora    & -        & -         & N/A    & 0.000 & 0.002 & 0.000   & 0.000 \\ \hline
Newspaper & -        & -         & -      & N/A   & 0.000 & 0.000   & 0.000 \\ \hline
NDR       & -        & -         & -      & -     & N/A   & 0.000   & 0.000 \\ \hline
CHILDES   & -        & -         & -      & -     & -     & N/A     & 0.046 \\ \hline
Music     & -        & -         & -      & -     & -     & -       & N/A   \\ \hline
\end{tabular}
\end{table}

\newpage 

\section*{Appendix F: Statistical Test Results for Language Pairs} 
The tables in this section list the p-values from the statistical test
of the Brunner-Munzel method for every pair of languages with at least
10 samples. \tabref{test-p-taylor-lang} and
\tabref{test-p-ebeling-lang} are for words, whereas
\tabref{test-p-ebeling-char-lang} is for characters. Because Taylor's
method is not applicable to scripts with a small number of characters,
no table is given for that case.
 
\begin{table}[h]
\centering
\caption{p-values of the Brunner-Manzel test for the Taylor exponent and language (words).}
\label{test-p-taylor-lang}
\tiny
\begin{tabular}{|l|c|c|c|c|c|c|c|c|c|c|c|c|}
\hline
         & Arabic & Hebrew & Finnish & Greek & French & Italian & Spanish & English & German & Dutch & Chinese & Japanese \\ \hline
Arabic   & N/A    & 0.000  & 0.000   & 0.002 & 0.000  & 0.032   & 0.013   & 0.000   & 0.587  & 0.000 & 0.000   & 0.173    \\ \hline
Hebrew   & -      & N/A    & 0.000   & 0.542 & 0.057  & 0.111   & 0.285   & 0.018   & 0.041  & 0.008 & 0.000   & 0.051    \\ \hline
Finnish  & -      & -      & N/A     & 0.004 &        & 0.404   & 0.000   & 0.000   & 0.000  & 0.000 & 0.000   & 0.000    \\ \hline
Greek    & -      & -      & -       & N/A   & 0.283  & 0.307   & 0.806   & 0.579   & 0.238  & 0.295 & 0.000   & 0.004    \\ \hline
French   & -      & -      & -       & -     & N/A    & 0.259   & 0.864   & 0.000   & 0.013  & 0.498 & 0.000   & 0.000    \\ \hline
Italian  & -      & -      & -       & -     & -      & N/A     & 0.204   & 0.095   & 0.034  & 0.177 & 0.008   & 0.013    \\ \hline
Spanish  & -      & -      & -       & -     & -      & -       & N/A     & 0.316   & 0.099  & 0.702 & 0.000   & 0.003    \\ \hline
English  & -      & -      & -       & -     & -      & -       & -       & N/A     & 0.226  & 0.021 & 0.000   & 0.002    \\ \hline
German   & -      & -      & -       & -     & -      & -       & -       & -       & N/A    & 0.025 & 0.079   & 0.477    \\ \hline
Dutch    & -      & -      & -       & -     & -      & -       & -       & -       & -      & N/A   & 0.000   & 0.000    \\ \hline
Chinese  & -      & -      & -       & -     & -      & -       & -       & -       & -      & -     & N/A     & 0.350    \\ \hline
Japanese & -      & -      & -       & -     & -      & -       & -       & -       & -      & -     & -       & N/A      \\ \hline
\end{tabular}
\end{table}

\begin{table}[h]
\centering
\caption{p-values of the Brunner-Manzel test for the EN exponent and language (words).}
\label{test-p-ebeling-lang}
\tiny
\begin{tabular}{|l|c|c|c|c|c|c|c|c|c|c|c|c|}
\hline
         & Arabic & Hebrew & Finish & Greek & French   & Italian & Spanish & English & German & Dutch & Chinese & Japanese \\ \hline
Arabic   & N/A    & 0.000  & 0.321  & 0.634 & 0.296 & 0.013   & 0.000   & 0.000   & 0.027  & 0.001 & 0.000   & 0.000    \\ \hline
Hebrew   & -      & N/A    & 0.000  & 0.001 & 0.000    & 0.000   & 0.000     & 0.000   & 0.000  & 0.000 & 0.000   & 0.000    \\ \hline
Finnish  & -      & -      & N/A    & 0.423 & 0.092    & 0.006   & 0.000   & 0.000   & 0.008  & 0.001 & 0.000   & 0.000    \\ \hline
Greek    & -      & -      & -      & N/A   & 0.723    & 0.046   & 0.012   & 0.001   & 0.035  & 0.141 & 0.032   & 0.000    \\ \hline
French   & -      & -      & -      & -     & N/A      & 0.056   & 0.000   & 0.000   & 0.064  & 0.022 & 0.004   & 0.000    \\ \hline
Italian  & -      & -      & -      & -     & -        & N/A     & 0.691   & 0.320   & 0.554  & 0.427 & 0.826   & 0.193    \\ \hline
Spanish  & -      & -      & -      & -     & -        & -       & N/A     & 0.392   & 0.827  & 0.112 & 0.677   & 0.146    \\ \hline
English  & -      & -      & -      & -     & -        & -       & -       & N/A     & 0.902  & 0.002 & 0.031   & 0.155    \\ \hline
German   & -      & -      & -      & -     & -        & -       & -       & -       & N/A    & 0.267 & 0.421   & 0.984    \\ \hline
Dutch    & -      & -      & -      & -     & -        & -       & -       & -       & -      & N/A   & 0.354   & 0.000    \\ \hline
Chinese  & -      & -      & -      & -     & -        & -       & -       & -       & -      & -     & N/A     & 0.037    \\ \hline
Japanese & -      & -      & -      & -     & -        & -       & -       & -       & -      & -     & -       & N/A      \\ \hline
\end{tabular}
\end{table}

\begin{table}[h]
\centering
\caption{p-values of the Brunner-Manzel test for the EN exponent and language (characters). }
\label{test-p-ebeling-char-lang}
\tiny
\begin{tabular}{|l|c|c|c|c|c|c|c|c|c|c|c|c|}
\hline
         & Arabic & Hebrew & Finnish & Greek & French & Italian & Spanish & English & German & Dutch & Chinese & Japanese \\ \hline
Arabic   & N/A    & 0.000  & 0.558   & 0.118 & 0.268  & 0.110   & 0.002   & 0.000   & 0.000  & 0.000 & 0.000   & 0.000    \\ \hline
Hebrew   & -      & N/A    & 0.000   & 0.000 & 0.000  & 0.000   & 0.000   & 0.000   & 0.000  & 0.000 & 0.000   & 0.000    \\ \hline
Finnish  & -      & -      & N/A     & 0.165 & 0.759  & 0.339   & 0.007   & 0.000   & 0.000  & 0.015 & 0.002   & 0.000    \\ \hline
Greek    & -      & -      & -       & N/A   & 0.323  & 0.955   & 0.169   & 0.087   & 0.023  & 0.594 & 0.177   & 0.046    \\ \hline
French   & -      & -      & -       & -     & N/A    & 0.232   & 0.014   & 0.000   & 0.000  & 0.030 & 0.000   & 0.000    \\ \hline
Italian  & -      & -      & -       & -     & -      & N/A     & 0.497   & 0.312   & 0.086  & 0.699 & 0.792   & 0.291    \\ \hline
Spanish  & -      & -      & -       & -     & -      & -       & N/A     & 0.857   & 0.130  & 0.334 & 0.736   & 0.901    \\ \hline
English  & -      & -      & -       & -     & -      & -       & -       & N/A     & 0.142  & 0.117 & 0.490   & 0.588    \\ \hline
German   & -      & -      & -       & -     & -      & -       & -       & -       & N/A    & 0.038 & 0.077   & 0.175    \\ \hline
Dutch    & -      & -      & -       & -     & -      & -       & -       & -       & -      & N/A   & 0.374   & 0.062    \\ \hline
Chinese  & -      & -      & -       & -     & -      & -       & -       & -       & -      & -     & N/A     & 0.923    \\ \hline
Japanese & -      & -      & -       & -     & -      & -       & -       & -       & -      & -     & -       & N/A      \\ \hline
\end{tabular}
\end{table}

 
\end{document}